\documentclass[runningheads]{llncs}
\usepackage{graphicx}
\usepackage{tikz}
\usepackage{comment}
\usepackage{amsmath,amssymb} % define this before the line numbering.
\usepackage{color}

\usepackage{adjustbox}
\usepackage{multirow}
\usepackage{threeparttable}
\usepackage[linesnumbered, ruled]{algorithm2e}
\usepackage{mathrsfs,amssymb}
\usepackage{bm}
\usepackage{cite}
\usepackage{subfig}
\usepackage{makecell}

\usepackage{array}
\newcommand{\PreserveBackslash}[1]{\let\temp=\\#1\let\\=\temp}
\newcolumntype{C}[1]{>{\PreserveBackslash\centering}p{#1}}

\usepackage{floatrow}
\newfloatcommand{capbtabbox}{table}[][\FBwidth]

\begin{document}
% \linenumbers
\pagestyle{headings}
\mainmatter
\def\ECCVSubNumber{6451}  % Insert your submission number here

\title{Video Anomaly Detection by Solving Decoupled Spatio-Temporal Jigsaw Puzzles} % Replace with your title\

\begin{comment}
\titlerunning{ECCV-22 submission ID \ECCVSubNumber} 
\authorrunning{ECCV-22 submission ID \ECCVSubNumber} 
\author{Anonymous ECCV submission}
\institute{Paper ID \ECCVSubNumber}
\end{comment}
%******************

% \begin{comment}
\titlerunning{VAD by Solving Decoupled Spatio-Temporal Jigsaw Puzzles}
\author{Guodong Wang\inst{1,2} \and
Yunhong Wang\inst{2} \and
Jie Qin\inst{3} \and
Dongming Zhang\inst{4} \and \\
Xiuguo Bao\inst{4} \and
Di Huang\inst{1,2}$\thanks{Corresponding author (ORCID: 0000-0002-2412-9330).}$}

\authorrunning{G. Wang et al.}
\institute{SKLSDE, Beihang University, Beijing, China \and
SCSE, Beihang University, Beijing, China \and
CCST, NUAA, Nanjing, China \and
CNCERT/CC,
Beijing, China \\
\email{\{wanggd,yhwang,dhuang\}@buaa.edu.cn},
\email{qinjiebuaa@gmail.com}, \\
\email{zhdm@cert.org.cn},
\email{baoxiuguo@139.com}}
% \end{comment}
%******************
\maketitle

\begin{abstract}

    Video Anomaly Detection (VAD) is an important topic in computer vision. Motivated by the recent advances in self-supervised learning, this paper addresses VAD by solving an intuitive yet challenging pretext task, \emph{i.e.}, spatio-temporal jigsaw puzzles, which is cast as a multi-label fine-grained classification problem. Our method exhibits several advantages over existing works: 1) the spatio-temporal jigsaw puzzles are decoupled in terms of spatial and temporal dimensions, responsible for capturing highly discriminative appearance and motion features, respectively; 2) full permutations are used to provide abundant jigsaw puzzles covering various difficulty levels, allowing the network to distinguish subtle spatio-temporal differences between normal and abnormal events; and 3) the pretext task is tackled in an end-to-end manner without relying on any pre-trained models. Our method outperforms state-of-the-art counterparts on three public benchmarks. Especially on ShanghaiTech Campus, the result is superior to reconstruction and prediction-based methods by a large margin.

\keywords{video anomaly detection; spatio-temporal jigsaw puzzles; multi-label classification}
\end{abstract}

\section{Introduction}

\label{sec:intro}

Video anomaly detection (VAD) refers to the task of detecting unexpected events that deviate from the normal patterns of familiar ones. Recently, it has become a very important task in the community of computer vision and pattern recognition with the exponential increase of video data captured from various scenarios. VAD is rather challenging as abnormal events are infrequent in real world and unbounded in category, jointly making typical supervised methods inapplicable due to the unavailability of balanced normal and abnormal samples for training. Therefore, VAD is generally performed in a one-class learning manner where only normal data are given \cite{liu2018future, liu2021hybrid, georgescu2021anomaly}.

\begin{figure*}[!h]
	\centering
	\includegraphics[width=1.0\textwidth]{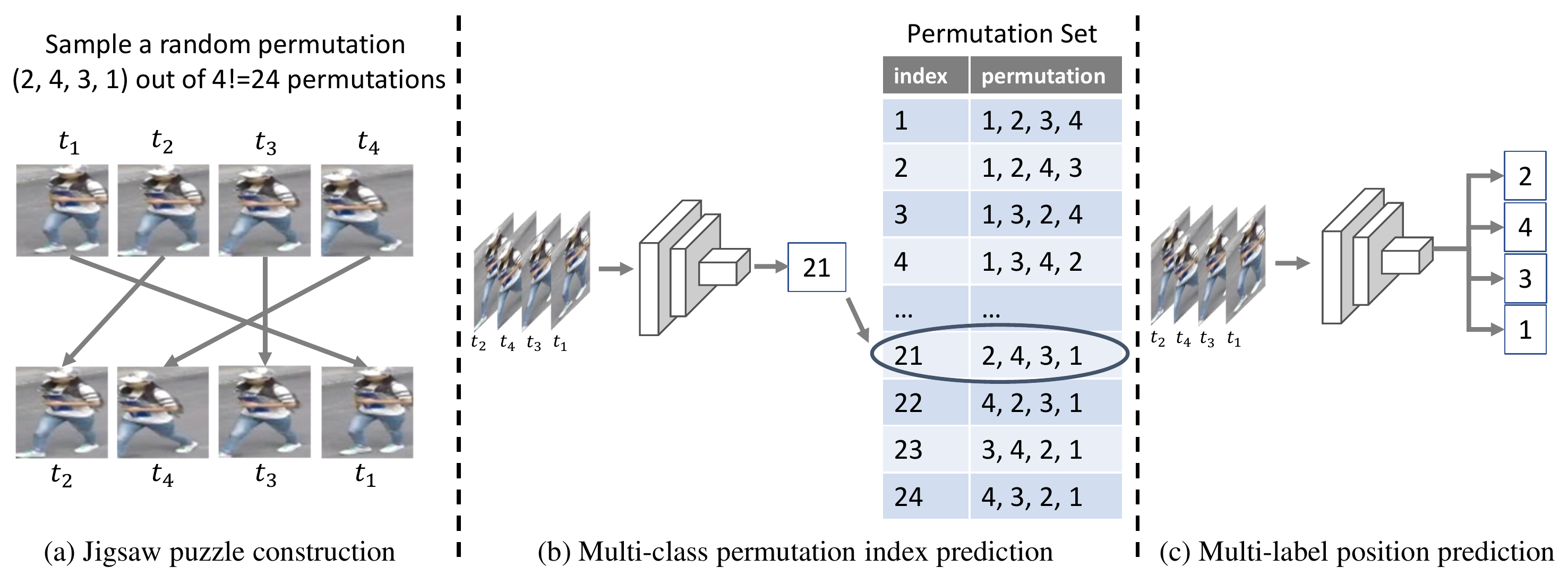}
	\caption{Multi-class index classification \emph{vs.} multi-label position classification. (a) Jigsaw puzzle construction. We permute the original sequence based on a randomly selected permutation from all possible ones. (b) Multi-class permutation index prediction. Traditional methods \cite{noroozi2016unsupervised, lee2017unsupervised, kim2019self} take a permutation as one class out of 4!=24 classes. (c) Multi-label permutation position prediction (\textbf{ours}). We directly output multiple predictions, indicating the absolute position in the original sequence for each frame.}
	\label{fig:tease}
\end{figure*}

In this regime, a series of VAD approaches \cite{liu2018future, feng2021Conv, luo2017remembering, gong2019memorizing, MM21, yu2020cloze, lee2019bman, nguyen2019anomaly} have been proposed, among which reconstruction and prediction based methods are two representative paradigms in the context of deep learning. Reconstruction based methods \cite{gong2019memorizing, nguyen2019anomaly} build models, \emph{e.g.}, autoencoders and generative adversarial networks (GAN), to recover input frames, and examples with high reconstruction errors are identified as anomalies at test time. In consideration of the temporal coherence, prediction based methods render missing frames \emph{e.g.}, middle frames \cite{yu2020cloze} or future frames \cite{liu2018future, feng2021Conv}, according to motion continuity. The difference between the predicted frame and its corresponding ground-truth suggests the probability of anomaly occurring.

The two types of methods above report promising performance; however, as stated in \cite{gong2019memorizing, munawar2017limiting, zong2018deep}, they aim at high-quality pixel generation, and even though the networks are only trained to perfectly match normal examples, their inherent generalization abilities still make the anomalies well reconstructed or predicted, especially for static objects, \emph{e.g.}, a stopped car in a pedestrian area. To address this, some follow-up studies attempt to boost the accuracy through incorporating memory modules \cite{gong2019memorizing, park2020learning}, modeling optical flows \cite{liu2018future}, redesigning specific architectures \cite{feng2021Conv}, \emph{etc}.

More recently, self-supervised learning has opened another avenue for VAD with significantly improved results. Different from the unsupervised generative solutions, self-supervised learning based methods explore supervisory signals for learning representations from unlabeled data \cite{wang2020cluster, georgescu2021anomaly}, and current investigations mainly differ in the design of pretext tasks. Wang \emph{et al.} \cite{wang2020cluster} propose an instance discrimination task to establish subcategories of normality as clusters and examples far away from the cluster centers are determined as anomalies. Georgescu \emph{et al.} \cite{georgescu2021anomaly} deliver an advancement by a model jointly considering multiple pretext tasks including discriminating arrow of time and motion regularity, middle frame reconstruction and knowledge distillation. Nevertheless, their pretext tasks are basically defined as binary classification problems, making them not so competent at learning highly discriminative features to distinguish \emph{subtle spatio-temporal differences} between normal and abnormal events. Additionally, these methods \cite{georgescu2021anomaly, wang2020cluster} depend on the networks pre-trained on large-scale datasets, \emph{e.g.}, ImageNet \cite{russakovsky2015imagenet} and Kinetics-400 \cite{kay2017kinetics}. % which are not always available in practical applications.

To circumvent the shortcomings aforementioned, in this paper, we propose a simple yet effective self-supervised learning method for VAD, through tackling an intuitive but challenging pretext task, \emph{i.e.}, spatio-temporal jigsaw puzzles. We hypothesize that successfully solving such puzzles requires the network to understand the very detailed spatial and temporal coherence of video frames by learning powerful spatio-temporal representations, which are critical to VAD. To this end, we take into account full possible permutations, rather than a subset produced by a heuristic permutation selection algorithm \cite{noroozi2016unsupervised}, to increase the difficulty of jigsaw puzzles with the aim of offering fine-grained supervisory signals for discriminative features. Based on the observation that anomalous events usually involve abnormal appearances and abnormal motions, we decouple spatio-temporal jigsaw puzzles in terms of spatial and temporal dimensions, responsible for modeling appearance and motion patterns, respectively, which technically facilitates optimization compared to solving 3D jigsaw puzzles \cite{ahsan2019video}. To be specific, we first randomly select a permutation from $n!$ possible permutations, where $n$ is the number of elements in the sequence. With this permutation, we then spatially shuffle patches within frames to construct spatial jigsaw puzzles or temporally shuffle a sequence of consecutive frames to build temporal ones. The training objective is to recover an original sequence from its spatially or temporally permuted version. Unlike existing methods for learning general visual representations \cite{noroozi2016unsupervised, lee2017unsupervised, kim2019self} which treat jigsaw puzzle solving as a multi-class classification problem where each permutation corresponds to a class (Figure \ref{fig:tease} (b)), we cast it as a multi-label learning problem (Figure \ref{fig:tease} (c)), allowing the method to be extendable to more advanced jigsaw puzzles with more pieces and free from significantly increased memory consumption. During inference, the confidence of the prediction with respect to unshuffled frames or images serves as the regularity score for anomaly detection. Abnormal events are expected to have lower confidence scores because they are unseen in training.

Compared to prior work on self-supervised VAD \cite{georgescu2021anomaly, wang2020cluster}, the advantages of our method are three-fold. \textbf{First}, we dramatically simplify the self-supervised learning framework by solving only a single pretext task, which is decoupled into the spatial and temporal jigsaw puzzles, corresponding to modeling normal appearance and motion patterns, respectively. \textbf{Second}, full possible permutations are employed to produce large-scale learning samples of a high diversity, allowing the network to capture subtle spatio-temporal anomalies from the pretext task. To ensure computational efficiency, we formulate puzzle solving as a multi-label learning problem, accommodating a factorial of number of variations. \textbf{Third}, our method is free from any pre-trained networks, because solving the challenging pretext itself helps to learn rich and discriminative spatio-temporal representations. It achieves state-of-the-art results on three public benchmarks, especially on the ShanghaiTech Campus dataset \cite{luo2017revisit}.

\section{Related Work}

\textbf{Video anomaly detection.} While early studies \cite{adam2008robust, antic2011video, cong2013abnormal} advocate manually-designed appearance and motion features for VAD, recent methods leverage the powerful representation capabilities of deep neural networks to automatically learn features from video events and deliver better performance. Most VAD methods follow the way of per-pixel generation, with reconstruction based and prediction based ones being the two important lines. Reconstruction based methods \cite{hasan2016learning, luo2017remembering, fan2020video} learn to recover input frames or clips, while prediction based methods learn to predict missing frames, such as future frame prediction \cite{lu2019future, liu2018future, MM21} or middle frame completion \cite{yu2020cloze, lee2019bman}. The combination of reconstruction and prediction as a hybrid solution is also explored in \cite{ye2019anopcn, zhao2017spatio, morais2019learning}.
These methods aim at high-quality pixel generation during training and examples with large reconstruction or prediction errors are identified as anomalies.
However, these networks often exhibit strong generalization abilities on anomalies (even though they are unseen in training), leading to decent reconstruction or prediction quality. The use of memory mechanisms \cite{gong2019memorizing, park2020learning} or multi-modal data (\emph{e.g.}, optical flows \cite{liu2021hybrid, nguyen2019anomaly} and RGB differences \cite{chang2020clustering}) suppresses the generalization ability to some extent but the improvement is far from perfect given the additional computation and memory consumption.

\begin{figure*}[!h]
	\centering
	\includegraphics[width=0.9\textwidth]{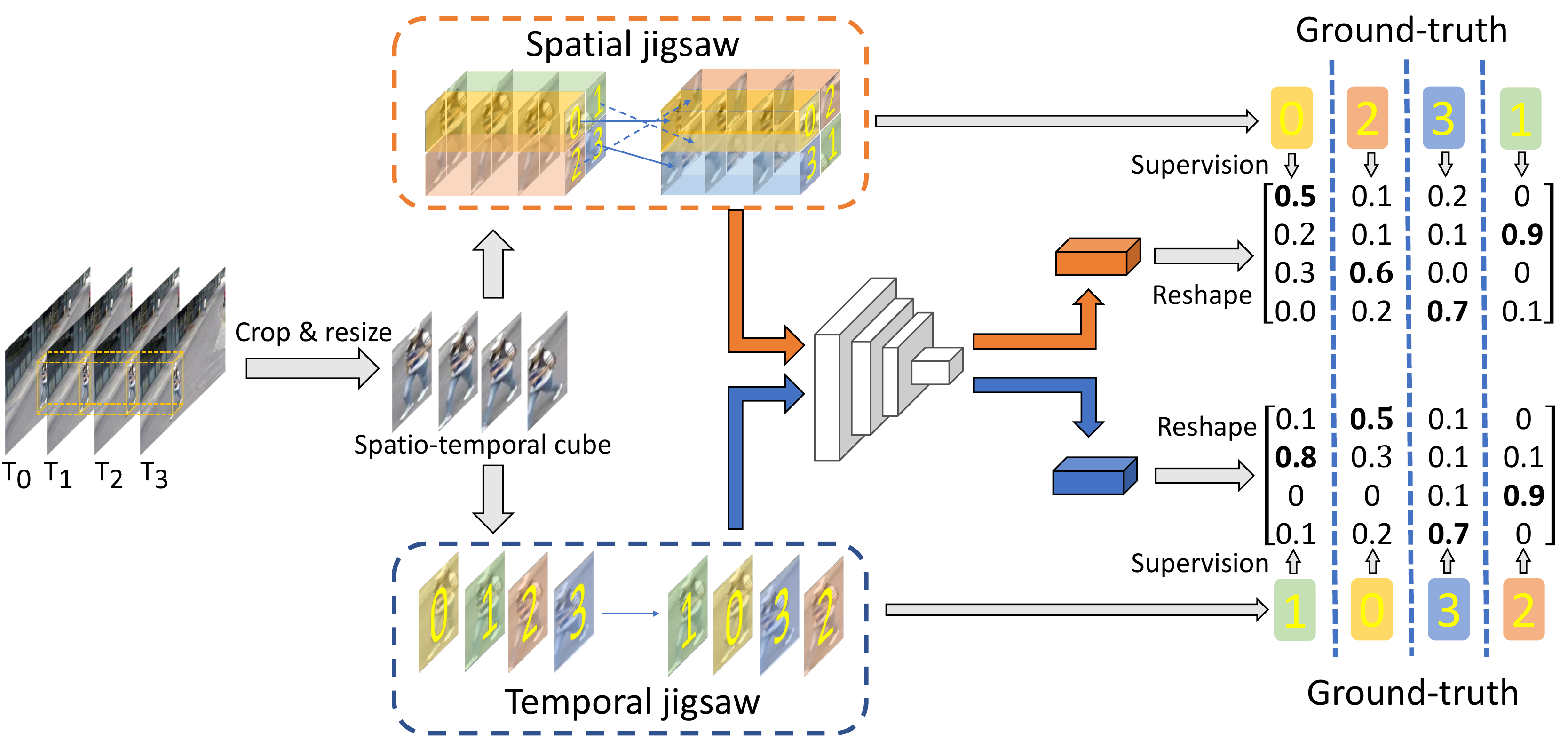}
	\caption{Method overview. We devise a pretext task including temporal and spatial jigsaw puzzles, for self-supervised learning spatio-temporal representations. Based on the object-centric spatio-temporal cubes, we create jigsaw puzzles by performing temporal and spatial shuffling. The network comprises a shared 3D convolution backbone followed by two disjoint heads to predict the permutation used for shuffling frames in time and patches in space, respectively. Each column of a matrix denotes the prediction of an entry in the permutation.}
	\label{fig:overall_arch}
\end{figure*}

\textbf{Self-supervised learning.} Self-supervised learning (SSL) is a generic learning framework which seeks supervisory signals from data only. It can be broadly categorized into constructing pretext tasks and conducting constrastive learning. \textbf{Pretext tasks.} For images, pretext tasks typically include solving jigsaw puzzles \cite{noroozi2016unsupervised}, coloring images \cite{zhang2016colorful}, and predicting relative patches \cite{doersch2015unsupervised} or image rotations \cite{gidaris2018unsupervised}, \emph{etc}. For videos, a series of methods additionally exploit temporal information exclusive to videos based on verifying correct frame order \cite{misra2016shuffle}, sorting frame order \cite{lee2017unsupervised} or clip order \cite{xu2019self}, predicting playback speed \cite{benaim2020speednet} or arrow of time \cite{wei2018learning, pickup2014seeing}, \emph{etc}. Among the pretext tasks, jigsaw puzzle is widely explored and proves effective in learning visual representation, but related methods fail to leverage all possible permutations, which scales factorially with the input length. Cruz \emph{et al.} \cite{santa2018visual} avoid such a factorial complexity by directly predicting the permutation matrix that shuffles the original data. \textbf{Contrastive learning.} It is another prevalent self-supervised learning paradigm in which each instance is regarded as a category. Motivated by the success of self-supervised image representation learning such as SimCLR \cite{chen2020simple} and MoCo \cite{he2020momentum}, many extensions \cite{pan2021videomoco, lorre2020temporal} of constrastive learning are proposed to adapt image-based methods to the video domain. 

\textbf{SSL in VAD.} While these self-supervised methods prove very effective in generic representation learning, benefiting bundles of downstream recognition and detection tasks \cite{chen2020simple, he2020momentum, pan2021videomoco, lorre2020temporal}, the efforts that exploit SSL for VAD are very few. SSL based VAD methods capture spatio-temporal representations by either conducting constrastive learning or solving pretext tasks, diverging from per-pixel reconstruction or prediction based methods. Wang \emph{et al.} \cite{wang2020cluster} learn spatio-temporal representations via a contrastive learning framework with a cluster attention mechanism. However, it requires a large number of training samples and customized data augmentation strategies. Georgescu~\emph{et al.} \cite{georgescu2021anomaly} train a 3D convolutional neural network jointly on three self-supervised proxy tasks and knowledge distillation for VAD. These proxy tasks are easy to solve, preventing the network from learning highly discriminative representations for VAD.

In this work, we design a more challenging pretext task, \emph{i.e.}, solving spatio-temporal jigsaw puzzles. Though solving jigsaw puzzles as a pretext task has been investigated for SSL by shuffling spatial layout \cite{noroozi2016unsupervised}, temporal order \cite{lee2017unsupervised} or their combination \cite{kim2019self}, they mostly formulate it as a multi-class classification task, therein each type of permutation corresponds to one class. The representations learned from the pretext tasks prove very effective evaluated in a series of downstream tasks, \emph{e.g.}, image retrieval and action recognition; however, its potential on VAD remains unexplored. A straightforward solution is directly applying these methods \cite{noroozi2016unsupervised, lee2017unsupervised} to learn representations of normal events for VAD. Nevertheless, this simple adaptation is sub-optimal since they only focus on modeling either appearances or motions while abnormal appearances and abnormal motions intertwine with each other in anomalous events in videos. Kim \emph{et al.} \cite{kim2019self} manage to solve space-time cubic puzzles, however, the multi-class formulation restricts itself to more advanced jigsaw puzzles (in fact they only leverage four-piece jigsaw puzzles), leading to inferior performance, shown in Table \ref{tab:numer_of_perm}. Though Ahsan \emph{et al.} \cite{ahsan2019video} consider directly solving 3D spatio-temporal jigsaw puzzles, we empirically find that its performance is not as good as expected due to the extreme difficulty of solving 3D jigsaw puzzles. For example, a cube compressing 7 frames with $3\times 3$ grid results in $7! \times (3\times 3)!=1,828,915,200$ possibles permutations. Therefore, we decompose the 3D spatio-temporal jigsaw puzzles into spatial and temporal jigsaw puzzles, corresponding to learning appearance and motion patterns, respectively.

\section{Method}

\subsection{Overview}

Figure \ref{fig:overall_arch} shows the pipeline of the proposed method, in which a sequence of four frames is used as an example for easy illustration. The method contains three steps: object-centric cube extraction, puzzle construction, and puzzle solving. We first employ an off-the-shelf object detector \cite{redmon2018yolov3} to extract all objects in the frames and stack the objects along the time dimension to construct object-centric cubes. For each cube, we further apply spatial or temporal shuffling to construct the corresponding spatial or temporal jigsaw puzzle. Finally, a convolutional neural network, acting as a jigsaw solver, attempts to recover the original sequence from its spatially or temporally permuted version. The proposed method is equivalent to solving a multi-label classification problem and is trained in an end-to-end manner.

It is noteworthy that we \textbf{do not} use any optical flows or pre-trained models (except for the object detector). The spatial and temporal permutations are only applied in training, allowing fast inference with a single forward pass.

\subsection{Fine-grained Decoupled Jigsaw Puzzles}

In self-supervised learning, it is crucial to prepare neither ambiguous nor easy self-labeled data \cite{noroozi2016unsupervised}. Based on the observation \cite{noroozi2016unsupervised, lee2017unsupervised} that networks can learn richer spatio-temporal representations from a more difficult pretext, we introduce full permutations for fine-grained jigsaw puzzle construction with the aim of capturing subtle spatial and temporal differences.

We first extract a large number of objects of interest by applying a YOLOv3 detector \cite{redmon2018yolov3} pre-trained on MS-COCO \cite{lin2014microsoft} frame by frame, therein we only keep the localization information and discard the classification labels. For each object detected in the frame $i$, we construct an object-centric spatio-temporal cube by simply stacking patches cropped from its temporally adjacent frames $\{i - t, ..., i - 1, i, i + 1, ... i + t\}$ using the same bounding box and location. We rescale all the extracted patches into a fixed size, \emph{e.g.}, 64$\times$64. Based on the extracted object-centric spatio-temporal cubes, we prepare training samples by constructing spatial or temporal jigsaw puzzles.

\textbf{Spatial jigsaw.} Following \cite{noroozi2016unsupervised}, for each frame, we start by decomposing it into $n\times n$ equal-sized patches which are then randomly shuffled. We make all the frames in the cube share the same permutation meanwhile keep them in the chronological order. Different from \cite{noroozi2016unsupervised} that separately passes each patch into the network, we directly take as input the frames after spatial shuffling, which are of the same size with the original frame, \emph{i.e.}, $64\times 64$ in our setting.

\textbf{Temporal jigsaw.} To construct temporal jigsaw puzzles, we shuffle a sequence of $l$ frames without disorganizing the spatial content. Jenni~\emph{et al.} \cite{jenni2020video} reveal that the most effective pretext tasks for powerful video representation learning are those that can be solved by observing the largest number of frames. For instance, motion irregularity \cite{georgescu2021anomaly} can be easily detected by just comparing the first two frames, in contrast to observing the total number of frames to solve our temporal jigsaw puzzles, which is crucial for learning more discriminative representations of motion patterns. Note that we do not temporally shuffle the frame sequence containing only static contents since it is impossible to infer its temporal order by simply observing visual cues.

\subsection{Multi-label Supervision}
Our jigsaw solving task is essentially a permutation prediction problem. Recall that, to make the task more challenging for learning discriminative representations, we employ the full permutations to produce fine-grained jigsaw puzzles. Different from typical methods \cite{noroozi2016unsupervised, kim2019self, lee2017unsupervised} that formulate jigsaw puzzle solving as multi-class classification, therein each permutation is a class, we cast jigsaw puzzle solving as a multi-label classification problem and attempt to directly predict the absolute position of each frame or the location of each patch. 
For each frame in the temporally shuffled sequence, we predict the correct position in the original sequence, while for each patch in the spatially shuffled frame, we predict the correct location in the original splitting grid. The strategy reduces the complexity $\mathcal{O}(l!)$ to $\mathcal{O}(l^2)$, and it can thus be easily extended to input frames of longer sequences or finer grid-splits with negligible memory consumption.

We adopt the mixed training strategy where a training mini-batch consists of two disjoint sets: $Q_s$ and $Q_t$, denoting the sets of spatial and temporal jigsaw puzzles, respectively. Thus, the mini-batch has a total of $|Q_t| + |Q_s|$ samples. It is worth noting that the two solvers (heads) are only responsible for their own puzzle types, \emph{i.e.}, we do not rely on the temporal solver to deal with spatial jigsaw puzzles to avoid ambiguity and vice versa. Algorithm \ref{alg:alg1} provides more details for constructing puzzles in mini-batches.

We optimize the network using the cross-entropy (CE) loss. For a jigsaw puzzle $p$, its loss is computed as Eq.~(\ref{eq:loss}).
\begin{equation}
    L_p = \begin{cases} 
    \frac{1}{l} \sum_{i=1}^l CE(t_i, \hat{t_i}), &  p \in Q_t \\ 
    \frac{1}{n^2} \sum_{j=1}^{n^2} CE(s_j, \hat{s_j}), & p \in Q_s
    \end{cases},
    \label{eq:loss}
\end{equation}
where $t_i$ and $\hat{t_i}$ are the ground-truth and predicted positions of a frame in the original sequence, respectively, and $s_j$ and $\hat{s_i}$ are the ground-truth and predicted locations of a patch in the original splitting grid, respectively.

\begin{algorithm}
\caption{Puzzle construction in mini-batches}
\label{alg:alg1}
\LinesNumbered
\DontPrintSemicolon
\KwIn{object-centric spatio-temporal cubes $C$, ratio $r$ , frame length $l$,  number of patches $n^2$, threshold $\zeta$.}
\KwOut{sets of jigsaw puzzles: $Q_t, Q_s$.}
$Q_t \leftarrow \varnothing, Q_s \leftarrow \varnothing$\\
$P^t \leftarrow$ all permutations $[P^t_1, P^t_2, ..., P^t_{l!}]$\\
$P^s \leftarrow$ all permutations $[P^s_1, P^s_2, ..., P^s_{(n^2)!}]$\\
\For{c {\rm in} C}{
$p\leftarrow \mathcal{U}_{float}[0, 1]$     \tcp*{uniform sampling}
\eIf{$p \leq r$}{
        \eIf{$p \leq \zeta$}{
            $i \leftarrow 1$\
        }{
            $i \leftarrow \mathcal{U}_{int}[1, (n^2)!]$\\
        }
        $q \leftarrow {\rm SpatiallyShuffle}(c, P^s_i)$\\
        $Q_s \leftarrow Q_s \cup \{q\}$\\
    }{  
        $j \leftarrow  \mathcal{U}_{int}[1, l!]$\\ 
        $q \leftarrow {\rm TemporallyShuffle}(c, P^t_j)$\\
        $Q_t \leftarrow Q_t \cup \{q\}$\\
    }
}
\end{algorithm}

\subsection{VAD Inference}

Following the same protocol of object detection in training, for each object in frame $i$, we construct the corresponding object-centric cube by cropping the bounding boxes from its temporally adjacent frames $\{i - t, ..., i - 1, i, i + 1, ... i + t\}$. During inference, we reuse the built-in jigsaw solvers to obtain the regularity scores. We pass the object-centric cubes without performing spatial or temporal shuffling and obtain two matrices, $M_s$ and $M_t$, corresponding to spatial and temporal permutation predictions, respectively.

Intuitively, the diagonal entries of the matrices of normal events are larger than those of abnormal ones, as the network is only trained to recover the original normal sequences. We thus simply take the minimum prediction score of a sequence as its regularity, as in Eq. (\ref{eq:object_normality}).

\begin{align}
\left\{    
	\begin{aligned}
    r_s &= {\rm min} ({\rm diag}({M_s}))  \\ 
    r_t &= {\rm min} ({\rm diag}({M_t}))
	\end{aligned}
\right. ,
\label{eq:object_normality}
\end{align}

\noindent where $\rm diag(\cdot)$ extracts the matrix diagonal, $M_s$ and $M_t$ are predicted by the spatial or temporal jigsaw solver, and $r_s$ and $r_t$ indicate the object-level regularity scores, respectively. We select the minimum score along the diagonal of the matrix as the resulting object-level regularity score, since an example is likely anomalous as long as one frame or patch is wrongly predicted, in accordance with fine-grained multi-label supervision in training. Similarly, we obtain the frame-level regularity score $R_s$ ($R_d$) by simply selecting the minimum object-level regularity score in the frame. Similar to \cite{georgescu2021anomaly}, we also apply a 3D mean filter to create a smooth anomaly score map. Following \cite{liu2018future, feng2021Conv, ye2019anopcn}, we normalize the irregularity scores of all frames in each video:

\begin{align}
\left\{    
	\begin{aligned}
    R_s &= \frac{R_s - {\rm min}(R_s)}{{\rm max}(R_s) - {\rm min}(R_s)} \\ 
    R_t &= \frac{R_t - {\rm min}(R_t)}{{\rm max}(R_t) - {\rm min}(R_t)}
	\end{aligned}
\right. .
\label{eq:object_normality}
\end{align}

The final frame-level regularity score $R$ (Eq. (\ref{eq:final_score})) is the weighted average of $R_s$ and $R_t$, followed by a temporal 1-D Gaussian filter. 

\begin{equation}
R = w * R_s + (1 - w) * R_t. 
\label{eq:final_score}
\end{equation}

\section{Experiments}

\subsection{Datasets}

We present the experimental results on three popular benchmarks, namely UCSD Ped2 \cite{mahadevan2010anomaly}, CUHK Avenue \cite{lu2013abnormal}, ShanghaiTech Campus \cite{luo2017revisit}. 

\textbf{UCSD Ped2 \cite{mahadevan2010anomaly}.} Ped2 contains 16 training videos and 12 test videos captured by a fixed camera. Example objects are pedestrians, bikes, and vehicles. Each video has a resolution of 240$\times$360 pixels in gray scale.

\textbf{CUHK Avenue \cite{lu2013abnormal}.} Avenue consists of 16 training videos and 21 test videos, respectively. It includes a total number of 47 abnormal events with throwing bag and moving toward/away from the camera being example anomalies. Each video has a resolution of 360$\times$640 RGB pixels.

\textbf{ShanghaiTech Campus (STC) \cite{luo2017revisit}. } It contains 330 training videos and 107 test videos covering 13 different scenes, making it more challenging than the other two datasets. Example anomalous events include car invading and person chasing. Each video has a resolution of 480$\times$856 RGB pixels.

\subsection{Implementation Details}
Since we train our network only on the object-centric cubes, the first stage of our method is object detection. For fair comparison, we follow \cite{georgescu2021anomaly} to adopt the same implementation\footnote{https://github.com/wizyoung/YOLOv3\_TensorFlow} of YOLOv3 \cite{redmon2018yolov3} and use the same configurations to filter out the detected objects with low confidence. We set the confidence thresholds to 0.5, 0.8, and 0.8 for Ped2, Avenue and STC, respectively. The confidence thresholds are shared during training/test for each dataset. The input to the network is a tensor of $l \times 64 \times 64 \times 3$ where $l$ is the length of the sequence. The difficulty levels of spatial and temporal jigsaw puzzles are adjusted by varying $n$ and $l$, respectively. We obtain the optimal results with $l=9$ on STC and $l=7$ on Ped2 and Avenue, and $n=3$ for all the three datasets. We empirically set $ r=0.5$ ($|Q_t|=|Q_s|$) and $w=0.5$ throughout all the experiments, indicating the equivalent importance of spatial and temporal branches. Considering that Avenue and Ped2 are relatively small compared to the scale of spatial jigsaw puzzles (9!=362,800), we do not perform spatial shuffling with the probability $\zeta =1e-4$ (Line 7 in Algorithm \ref{alg:alg1}). 

Our framework is implemented using the PyTorch library \cite{paszke2017automatic} and trained in an end-to-end manner. We adopt the Adam optimizer with $\beta_1=0.9, \beta_2=0.999$. The learning rate is 1e-4. We train the network for 100 epochs on Avenue and STC and 50 epochs on UCSD Ped2, and set the batch size as 192.

\subsection{Evaluation Metric}
Following the widely-adopted evaluation metric used in VAD community \cite{liu2018future, wang2020cluster, liu2021hybrid, gong2019memorizing, MM21, yu2020cloze}, we report the frame-level area under the curve (AUC) of Receiver Operation Characteristic (ROC) with respect to the ground-truth annotations by varying the threshold. Specifically, we concatenate all the frames in dataset and then compute the overall frame-level AUC, \emph{i.e.}, micro-averaged AUROC \cite{georgescu2020background}.

\subsection{Experimental Results}

\begin{table}[!t]
	\centering
	\caption{Comparison with state-of-the-art methods in terms of micro-AUROC (\%). The best and second-best results are bold and underlined, respectively. $^*$ denotes that micro-AUROC is reported. SSL is short for self-supervised learning.}
	\begin{adjustbox}{width=0.7\textwidth}
	\begin{threeparttable}
    	\begin{tabular}{c|c|ccc}
    		\hline
    		\textbf{Type}     & \textbf{Method}   & \textbf{Ped2} & \textbf{Avenue} & \textbf{STC} \\ \hline\hline
    		
    		\multirow{8}{*}{\rotatebox{90}{reconstruction}}
    		& Conv-AE \cite{hasan2016learning}       & 90.0 & 70.2 & -    \\ 
    		& StackRNN \cite{luo2017revisit}              & 92.2 & 81.7 & 68.0  \\ 
    		& Mem-AE \cite{gong2019memorizing}      & 94.1 & 83.3 & 71.2 \\
    		& AM-Corr \cite{nguyen2019anomaly}          & 96.2 & 86.9 & -    \\
    		& MNAD-Recon. \cite{park2020learning}        & 90.2 & 82.8 & 69.8 \\
    		& ClusterAE \cite{chang2020clustering}  & 96.5 & 86.0 & 73.3  \\
    		& VEC \cite{yu2020cloze}                & 97.3 & 90.2 & 74.8  \\
    		& LNRA (Patch based) \cite{astrid2021learning} & 94.8 & 84.9 & 72.5 \\
    		& LNRA (Skip frame based) \cite{astrid2021learning} & 96.5 & 84.7 & 76.0 \\
    		& I3D-Recons.                           & -   & 69.3 - & 69.4 \\
    		
    		\hline\hline

    		\multirow{6}{*}{\rotatebox{90}{prediction}}
    		& Frame-Pred. \cite{liu2018future}      & 95.4 & 85.1 & 72.8 \\
    		& BMAN \cite{lee2019bman}               & 96.6 & 90.0 & 76.2 \\
    		& Multipath-Pred. \cite{wang2021robust} & 96.3 & 88.3 & 76.6 \\
    		& MNAD-Pred. \cite{park2020learning}    & 97.0 & 88.5 & 70.5 \\
    		& CT-D2GAN \cite{feng2021Conv}          & 97.2 & 85.9 & 77.7 \\
    		& Bi-Prediction \cite{chen2020anomaly}  & 96.6 & 87.8 & - \\
 
    		\hline\hline
    		
    		\multirow{5}{*}{\rotatebox{90}{hybrid}}
    		& ST-CAE \cite{zhao2017spatio}          & 91.2 & 80.9 & -   \\
    		& MPED-RNN \cite{morais2019learning}    & -    & -    & 73.4 \\
    		& AnoPCN \cite{ye2019anopcn}            & 96.8 & 86.2 & 73.6 \\
    		& IntegradAE \cite{tang2020integrating} & 96.3 & 85.1 & 73.0 \\
    		& HF$^2$-VAD \cite{liu2021hybrid}       & \textbf{99.3} & 91.1 & 76.2 \\
    		
    		\hline\hline

    		\multirow{4}{*}{\rotatebox{90}{others}}
    		& SCL \cite{lu2013abnormal}             & -    & 80.9 & -    \\
    		& DeepOC \cite{wu2019deep}              & 96.9 & 86.6 & -    \\
            & CAE-SVM$^*$ \cite{ionescu2019object}  & 94.3 & 87.4 & 78.7 \\
            & Scene-Aware \cite{sun2020scene}       & -    & 89.6 & 74.7 \\
    		
    		\hline\hline

    		\multirow{3}{*}{\rotatebox{90}{SSL}}
    		& CAC \cite{wang2020cluster}            & -    & 87.0 & 79.3 \\
    		& SS-MTL$^*$ \cite{georgescu2021anomaly}& 97.5 & \underline{91.5} & \underline{82.4} \\
    		& \textbf{Ours}          & \underline{99.0}    & \textbf{92.2}    & \textbf{84.3} \\
    		\hline

    	\end{tabular}
    
    \end{threeparttable}
	\end{adjustbox}
    \label{tab:overall}
	\label{tab1}
\end{table}

Table \ref{tab:overall} shows the comparison results with different types of state-of-the-art approaches, where we can observe that our method delivers very impressive performance on all the three benchmarks. On the challenging STC, our method outperforms reconstruction-based, prediction-based, and hybrid methods by significant margins. For example, our method achieves 84.3\% while the best accuracy of previous generative methods is 77.7\% by CT-D2GAN \cite{feng2021Conv}. This suggests the superiority of self-supervised learning which captures discriminative representations of normal events by solving pretext tasks, bypassing the requirement for per-pixel generation. Additionally, compared to the VAD methods \cite{georgescu2021anomaly, wang2020cluster} leveraging self-supervised learning, we still achieve the best performance, boosting the second-best method \cite{georgescu2021anomaly} by 1.9\%. We attribute it to the design of our challenging pretext task, \emph{i.e.}, solving fine-grained spatio-temporal jigsaw puzzles with full permutations, which helps to learn discriminative representations. Note that we only use the VAD training set to train our network and do not use an extra model for either knowledge distillation \cite{georgescu2021anomaly} or transfer learning \cite{wang2020cluster}. On Avenue and UCSD Ped2, we also deliver very competitive performance, indicating that our method is robust to datasets of different scales.

\section{Ablation Study}

To understand the factors that contribute to the anomaly detection performance, we conduct ablation studies on STC and Avenue considering four key factors that control the puzzle difficulty: a) number of permutations; b) number of frames/patches; c) types of puzzles; and d) other pretexts beyond jigsaw solving. We also discuss the reliance on the object detector.

\begin{table}
    \centering
    \caption{Results of various numbers of permutations on STC and Avenue in terms of AUROC (\%). T and S represent the number of permutations for temporal and spatial jigsaw puzzle construction, respectively. Here, $l=7$ and $n^2=9$.}
    \begin{tabular}{c|c|c|c|cc}
    \hline
    
    \textbf{Exp. ID} & \textbf{Method} & \textbf{T} & \textbf{S} & \textbf{Avenue} &  \textbf{STC} \\ 
    
    \hline\hline
    A1 &  \multirow{4}*{\makecell[c]{Baseline \\ (Multi-class)}} &  504 & 504 & 84.6 & 76.8  \\ 
    A2 & & 5040 & 504 & 85.3  & 77.6   \\ 
    A3 & & 504 & 5040 & 87.0  & 78.1  \\ 
    A4 & & 5040 & 5040 & 87.5  & 78.5 \\ \hline\hline
    A5 & \multirow{5}*{\makecell[c]{Ours \\ (Multi-label)}} & 504 & 504 & 87.1 & 79.7 \\ 
    A6 & & 5040 & 504 & 87.5  &  81.2 \\ 
    A7 & & 504 & 5040 &  88.6 &  79.8 \\ 
    A8 & & 5040 & 5040 & 89.5 & 82.0 \\
    A9 & & 5040 & 362880 & \textbf{92.2} & \textbf{83.2} \\
    \hline
    \end{tabular}
    
    \label{tab:numer_of_perm}
\end{table}

\textbf{Number of permutations.} We constrain the number of permutations used to construct puzzles by selecting the subsets of full permutations using a Hamming distance based selection algorithm \cite{noroozi2016unsupervised}. We first build a baseline that follows the typical solution for solving puzzles, which considers a permutation as a class. During inference, the regularity scores are the probabilities of the spatio-temporal cubes not being spatially or temporally permuted. From Table \ref{tab:numer_of_perm}, both the baseline method and our method achieve improved results with a large number of permutations for both spatial and temporal puzzles, since the networks need to capture more discriminative representations to perceive subtle differences among jigsaw puzzles. Moreover, our method consistently outperforms the baseline for the same number of permutations. One possible reason is that the baseline attempts to discriminate jigsaw puzzles by different permutations, while we aim to predict the correct position of each frame/patch in the permutation in a more detailed way. Moreover, for advanced puzzles with more pieces, the baseline model fails due to memory limitation. In contrast, thanks to the multi-label classification formulation, our method can handle finer-grained puzzles with full permutations in a memory-friendly way, achieving the best (A9).

\begin{table}
    \centering
    \caption{Results of different numbers of frames/patches on STC and Avenue in terms of AUROC (\%). $l$ and $n^2$ denote the number of frames in an object-centric cube and the number of patches in the frames, respectively.}
    \begin{tabular}{c|c|c|cc}
    \hline
    \textbf{Exp. ID} & \bm{$l$} & \bm{$n^2$} & \textbf{Avenue}  & \textbf{STC}  \\
    \hline\hline
    B1 & 5 & 4 & 89.7 &  79.3 \\
    B2 & 7 & 4 & 90.2 &  80.2 \\
    B3 & 7 & 9 & \textbf{92.2} &  83.2 \\
    B4 & 9 & 4 & 88.6 &  81.1 \\
    B5 & 9 & 9 & 89.2 & \textbf{84.3} \\
    B6 & 9 & 16 & 87.9 &  80.4 \\
    \hline
    \end{tabular}
    \label{tab:numer_of_frames}
\end{table}

\textbf{Number of frames/patches}. With full permutations considered for puzzle construction, we next examine the effects of the number of frames ($l$) in the temporal dimension and the number of patches ($n^2$) in the spatial dimension. We do not try a larger $l$ ($l > $  9) as the object would go beyond the boundary of the spatio-temporal cube. From Table \ref{tab:numer_of_frames}, we can observe a trend of performance improvement when we increase $l$ and $n^2$ in a certain range. The observation is consistent with human beings who need more efforts to solve puzzles with more pieces. However, when we increase $l$ and $n^2$ further, the performance deteriorates. The reason lies in that the network is difficult to optimize especially on the spatial jigsaw puzzles, as each patch is very small ($16 \times 16$ pixels for $n^2=16$ in our setting) and thus causes ambiguity. 

\textbf{Types of puzzles}. Finally, we investigate the effects of solving spatial and temporal jigsaw puzzles for VAD. To this end, we design four alternative configurations based on when and which jigsaw puzzles are activated. Here, we set $l=7$ and $l=9$ for Avenue and STC, respectively; $n^2=9$ for both. Our method benefits more from solving spatial and temporal jigsaw puzzles in the training phase. For example, C3 indicates simultaneously solving spatial and temporal jigsaw puzzles during training, while C1 represents solving temporal jigsaw puzzles only. We activate the temporal solver only during testing for C1 and C3. In other words, the only difference between C1 and C3 is the training goal, \emph{i.e.}, multi-task \emph{vs.} single-task. Our method clearly benefits from multi-task learning and achieves better performance 82.7\% \emph{vs.} 78.6\% on STC. However, when only one type of puzzle is activated either during training or testing, the results are always worse than our complete version, namely C5. The observations are intuitive that anomalous events are caused by abnormal appearances and/or motions. Therefore, it is beneficial to include both types of jigsaw puzzles to detect both types of anomalies.

\textbf{Other pretexts beyond jigsaw solving.} We design other alternative pretext tasks considering the spatial dimension (\emph{e.g.}, rotation prediction \cite{komodakis2018unsupervised} and translation prediction \cite{hendrycks2019using}) and the temporal dimension (\emph{e.g.}, arrow of time prediction \cite{wei2018learning} and temporal order verification \cite{misra2016shuffle}). Our method achieves the best performance in Table \ref{tab:pretext}. It gives evidence that a proper design of the pretext task enabling fine-grained discrimination is essential for VAD. Compared to other pretexts, ours sets a more challenging task which requires the model to perceive every patch within a frame and every frame within a clip.

\begin{table}[!ht]
	\centering
	\caption{Results of different pretexts for VAD on STC in terms of AUROC (\%).}
	\begin{threeparttable}
    	\begin{tabular}{c|c|c|c}
    		\hline
    		\textbf{Exp. ID} & \textbf{Spatial}   & \textbf{Temporal}  & \textbf{STC} \\ \hline\hline
            D1  &  Rotation  &  Arrow of time & 72.9  \\
            D2  &  Rotation  &  Temporal order verification & 74.8  \\
            D3  &  Translation  &  Arrow of time & 73.0  \\
            D4  &  Translation  &  Temporal order verification & 75.6  \\
            D5  &  Translation  & Jigsaw (ours)  & 81.1 \\
            D6  &  Jigsaw (ours)  & Temporal order verification  & 78.3 \\
    	    D7  &  Jigsaw (ours)  & Jigsaw (ours)  & \textbf{84.3} \\
    		\hline
    	\end{tabular}
    \end{threeparttable}
    \label{tab:pretext}
\end{table}

\textbf{Object detector.} Although we mainly focus on object-level anomaly detection, our method can also be applied at frame level. To this end, we remove the object detector and report the results on the RetroTrucks dataset \cite{haresh2020towards}. For fair comparison with \cite{haresh2020towards}, we train an I3D \cite{wang2018non} network to predict the absolute position of the original sequence, since we observe that activating the temporal branch only is sufficient. Our method achieves the best performance in Table \ref{tab:obj_det}, even outperforming \cite{haresh2020towards} that incorporates object interaction reasoning.

\begin{table*}
\begin{floatrow}
\capbtabbox{
	\begin{threeparttable}
         \begin{tabular}{c|cc|cc|cc}
            \hline
            \multirow{2}*{\textbf{Exp. ID}} & \multicolumn{2}{c|}{\textbf{Train}} & \multicolumn{2}{c|}{\textbf{Test}} & \multirow{2}*{\textbf{Avenue}} & \multirow{2}*{\textbf{STC}} \\
            
            \cline{2-4}  \cline{4-5}
            & \textbf{T} & \textbf{S} & \textbf{T} & \textbf{S}  & &  \\
            \hline\hline
            C1 & \checkmark & - & \checkmark &  - & 78.9 & 78.6 \\
            C2 & - & \checkmark & - &  \checkmark & 86.7 & 76.0 \\
            C3 &\checkmark & \checkmark & \checkmark & & 86.9 & 82.7 \\
            C4 & \checkmark & \checkmark &  & \checkmark & 89.0 & 79.8 \\
            C5 &\checkmark & \checkmark & \checkmark & \checkmark & \textbf{92.2} &  \textbf{84.3} \\
            \hline
            \end{tabular}
    \end{threeparttable}

}{
 \caption{Results of different jigsaw puzzles. T and S are short for ``temporal'' and ``spatial'', respectively.}
 \label{tab:type_puzzle}
}

\capbtabbox{
	\begin{threeparttable}
    	\begin{tabular}{c|c}
    		\hline
    		\textbf{Method} & \textbf{RetroTrucks}  \\ \hline\hline
            Frame-Pred. \cite{liu2018future}   & 60.6 \\
            Mem-AE \cite{gong2019memorizing}   & 63.6 \\
            I3D \cite{haresh2020towards}       & 71.2 \\
            I3D + GCN \cite{haresh2020towards} & 71.5 \\
            Ours                               & \textbf{72.8} \\
    		\hline
    	\end{tabular}
    \end{threeparttable}

}{
 \caption{Results on RetroTrucks in terms of AUROC (\%).}
 \label{tab:obj_det}
 \small
}
\end{floatrow}
\end{table*}

\section{Conclusion}
In this work, we present a simple yet effective self-supervised learning framework for VAD through solving a challenging pretext task, \emph{i.e.}, spatio-temporal jigsaw puzzles, which are decoupled into spatial and temporal jigsaw puzzles for easy optimization. We emphasize that a challenging pretext task is key to learning discriminative spatio-temporal representations. To this end, we perform full permutations to generate a rich set of spatial and temporal jigsaw puzzles with varying degrees of difficulty, which allows the network to discriminate subtle spatio-temporal differences between normal and abnormal events. We reformulate the pretext task as a multi-label fine-grained classification problem, which is addressed in an efficient and end-to-end manner. Experiments show that our method achieves state-of-the-art on three popular benchmarks.

\section*{Acknowledgment}
This work is partly supported by the National Natural Science Foundation of China (62022011, U20B2069), the Research Program of State Key Laboratory of Software Development Environment (SKLSDE-2021ZX-04), and the Fundamental Research Funds for the Central Universities.

\clearpage
\bibliographystyle{splncs04}
\bibliography{egbib}

\appendix

\onecolumn
\clearpage

\begin{center}{\bf {\Large Video Anomaly Detection by Solving Decoupled Spatio-Temporal Jigsaw Puzzles}}
\end{center}
\begin{center}{\bf {\Large (Supplementary Material)}}
\end{center}

The supplementary material provides: 

\begin{itemize}
	\item detailed configuration of the network architecture.
	
	\item comparison in terms of macro-averaged AUROC metric \cite{georgescu2020background}.
	
	\item running time analysis.
	
	\item visual results on UCSD Ped2 \cite{mahadevan2010anomaly}, CUHK Avenue \cite{lu2013abnormal} and ShanghaiTech Campus (STC) \cite{luo2017revisit}.
	
	\item multi-label regression loss \emph{vs.} multi-label classification loss for training.
	
	\item action recognition experiment on UCF-101 \cite{soomro2012ucf101} using linear probing evaluation.
	
\end{itemize}

\section{Network Architecture}

The detailed configuration of the network is presented in Table \ref{tab:net}. The network consists of a shared convolutional part and two independent heads. The shared convolutional neural network (CNN) consists of 3D convolutions (conv) to extract spatio-temporal representations and 2D conv to aggregate spatial representations, while the two individual heads are fully connected (fc) layers. The shared part consists of three 3D blocks and one 2D block. Each 3D block comprises two 3D convolutional layers with the filters of $3\times 3\times 3$ and  a 3D max-pooling layer. Each convolutional layer is followed by an instance normalization (IN) layer \cite{ulyanov2016instance}, a ReLU activation layer. We perform 3D max-pooling along the spatial dimension in the first two blocks while the last 3D max-pooling layer performs global temporal pooling. The 2D block consists of a 2D convolutional layer, followed by an IN layer, a ReLU activation layer, a 2D dropout layer, and a 2D max-pooling layer. Both heads share the same configuration with two fc layers. We employ IN layers in the network since spatial and temporal jigsaw puzzles are instance-specific and independent of each other. Generally, we adopt the similar architecture (expect for the normalization layer) with the ``deep+wide" 3D CNN in \cite{georgescu2021anomaly} for fair comparison.

\begin{table}
	\centering
	\caption{The detailed network architecture. Global temporal pooling is denoted by “:”. $n^2$ and $l$ denote the number of patches in space dimension and the number of frames in time dimension, respectively.}
	\begin{tabular}{|c|C{1.5cm}|C{1.5cm}|}
		\hline
		\multirow{10}{*}{\rotatebox{90}{3D}}
		
		&  \multicolumn{2}{c|}{$3\times3\times3$ conv 32}     \\
		&  \multicolumn{2}{c|}{$3\times3\times3$ conv 32}     \\
		&  \multicolumn{2}{c|}{$1\times2\times2$ max-pooling} \\     
		&  \multicolumn{2}{c|}{$3\times3\times3$ conv 64}     \\
		&  \multicolumn{2}{c|}{$3\times3\times3$ conv 64}    \\
		&  \multicolumn{2}{c|}{$1\times2\times2$ max-pooling} \\
		&  \multicolumn{2}{c|}{$3\times3\times3$ conv 64}     \\ 
		&  \multicolumn{2}{c|}{$3\times3\times3$ conv 64}     \\
		&  \multicolumn{2}{c|}{$:\times2\times2$ max-pooling} \\
		
		\hline
		\multirow{3}{*}{\rotatebox{90}{2D}} 
		
		&\multicolumn{2}{c|}{$3\times3$ conv}               \\
		&\multicolumn{2}{c|}{dropout}                       \\
		&\multicolumn{2}{c|}{$2\times2$ max-pooling}        \\ \hline
		\multirow{2}{*}{\rotatebox{90}{Head}}
		& 512 fc &  512 fc              \\
		&  $(n^2)^2$ fc &  $l^2$ fc       \\

		\hline
	\end{tabular}
	\label{tab:net}
\end{table}

\section{Macro-averaged AUROC Comparison}

We note that most of the existing works \cite{liu2018future, wang2020cluster, liu2021hybrid, gong2019memorizing, MM21, yu2020cloze} report the micro-averaged AUROC by concatenating all frames in the dataset then computing the score while some \cite{georgescu2021anomaly, ionescu2019object} report macro-average AUROC by first computing the AUROC for each video then averaging these scores. Note that we report the micro-averaged AUROC in our main paper by default. Here, we also report the macro-averaged AUROC in Table \ref{tab:macro}. Clearly, we also achieve the best performance.

\section{Running Time}
All experiments are conducted on an NVIDIA RTX 2080 Ti GPU and an Intel(R) Xeon(R) CPU E5-2650 v4 @ 2.20GHz. For object detection, the YOLOv3 model \cite{redmon2018yolov3} takes about 20 milliseconds (ms) per frame. In the anomaly detection phase, our lightweight model infers the anomaly scores in 3 ms. With all components considered, our method runs at 28 FPS with an average of 5 objects per frame while the running speed of HF$^2$-VAD is about 10 FPS. The run-time bottleneck of our framework principally lies in object detection and spatio-temporal cube construction.

\begin{table}[!ht]
	\centering
	\caption{Comparison with state-of-the-art methods on frame-level performance in terms of macro-averaged AUROC (\%). The best and second-best results are bold and underlined, respectively. $^*$ denotes the results taken from \cite{georgescu2020background}.}
	\begin{threeparttable}
		\begin{tabular}{c|c|ccc}
			\hline
			\textbf{Year}     & \textbf{Method}   & \textbf{Ped2} & \textbf{Avenue} & \textbf{STC} \\ \hline
			
			2018 & Frame-Pred.$^*$ \cite{liu2018future}  & 98.1 & 81.7 & 80.6 \\
			\hline
			
			2019 & CAE-SVM$^*$ \cite{ionescu2019object}  & 97.8 & 90.4 & 84.9 \\ \hline
			2021 & SS-MTL \cite{georgescu2021anomaly}& \underline{99.8} & \underline{91.9} & \underline{89.3} \\ \hline
			% & \textbf{Ours} - light                 & -    & -    & - \\
			2022 & \textbf{Ours}   & \textbf{99.9}  & \textbf{93.0}  & \textbf{90.6} \\
			\hline

		\end{tabular}
		
	\end{threeparttable}
	\label{tab:macro}
\end{table}

\begin{figure}[!h]
	\centering
	\subfloat[Test video 02 from UCSD Ped2.]{
		\includegraphics[width=0.49\textwidth]{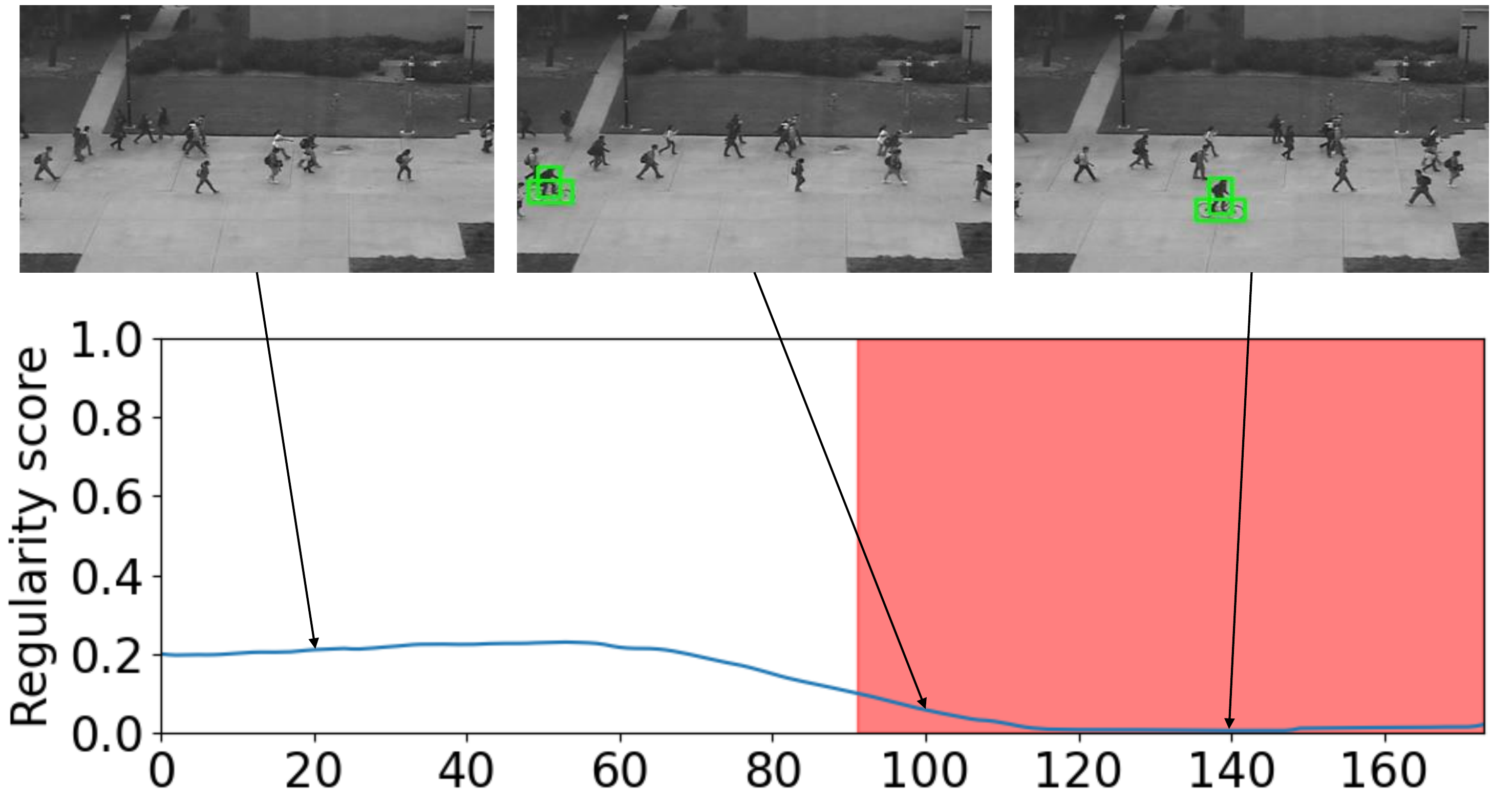}} 
	\subfloat[Test video 07 from UCSD Ped2.]{
		\includegraphics[width=0.49\textwidth]{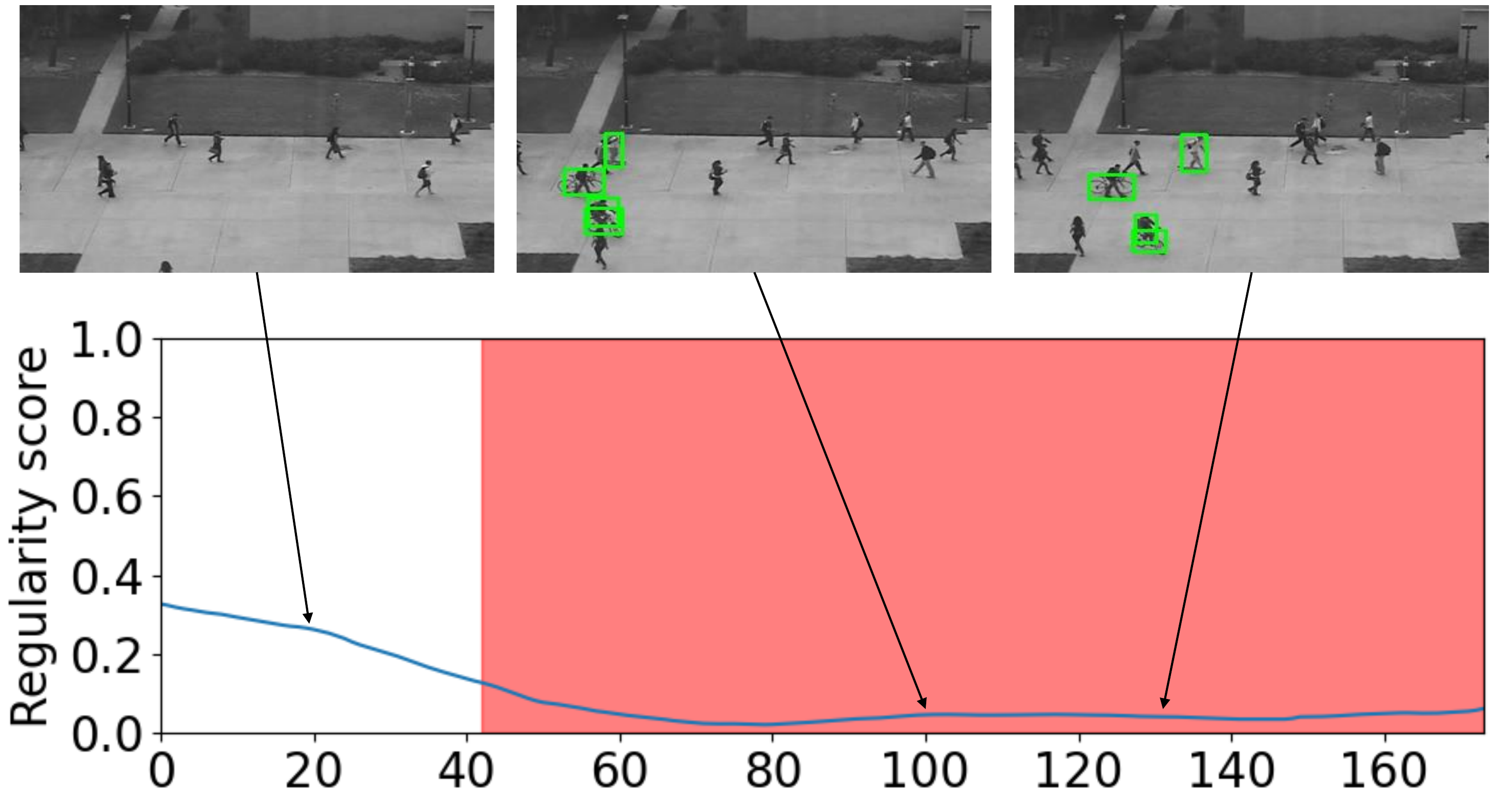}}
	
	\caption{Regularity score curves by our method on UCSD Ped2. The light red shaded regions represent the ground-truth segments of abnormal events.}
	\label{fig:ped2}
\end{figure}

\begin{figure}[!h]
	\centering
	\subfloat[Test video 16 from CUHK Avenue.]{
		\includegraphics[width=0.49\textwidth]{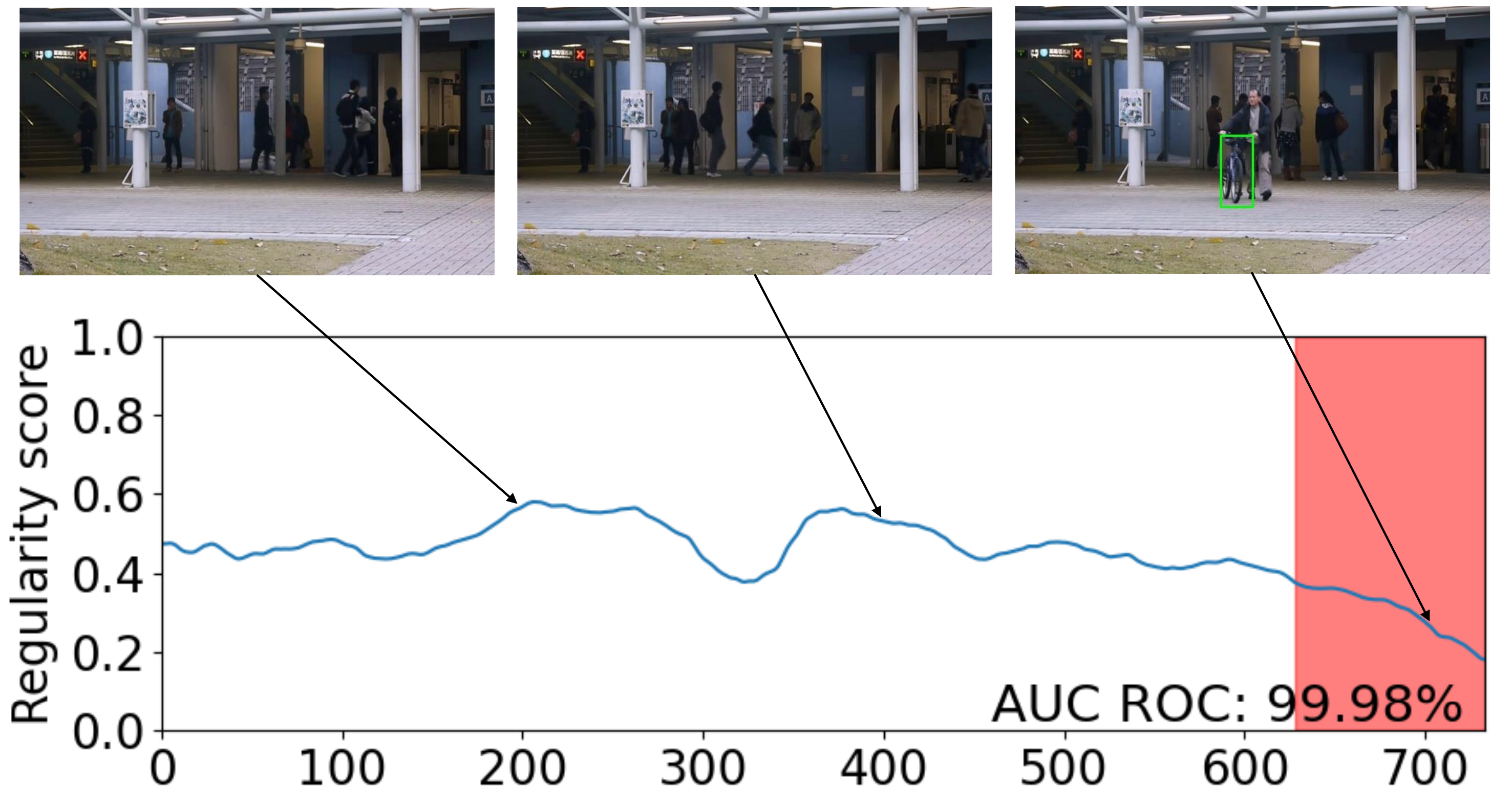}}
	\subfloat[Test video 02 from CUHK Avenue.]{
		\includegraphics[width=0.49\textwidth]{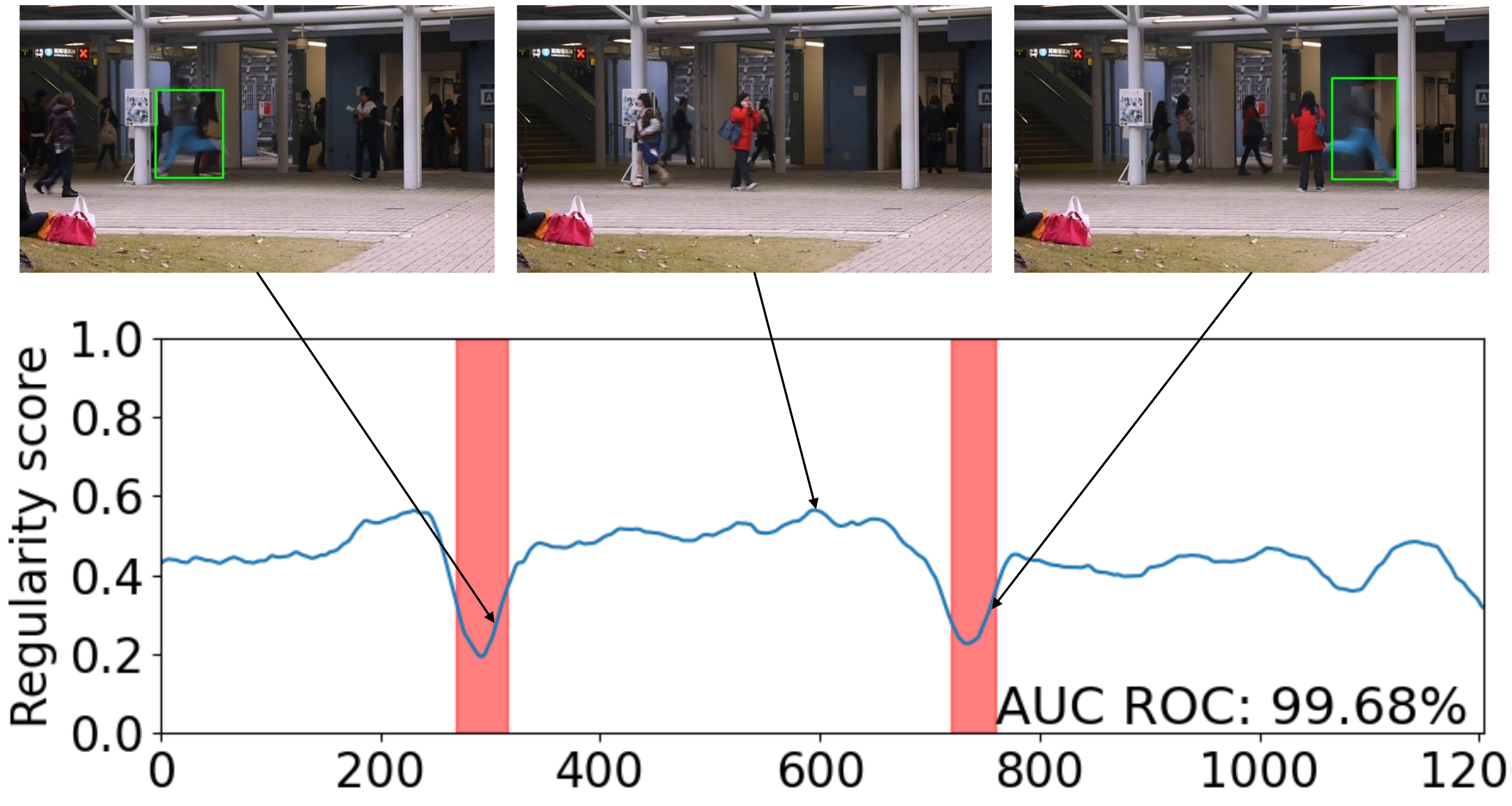}}
	
	\caption{Regularity score curves by our method on CUHK Avenue. The light red shaded regions represent the ground-truth segments of abnormal events.} 
	\label{fig:avenue}
\end{figure}

\begin{figure}[!h]
	\centering
	\subfloat[Test video 03\_0031 from STC.]{
		\includegraphics[width=0.49\textwidth]{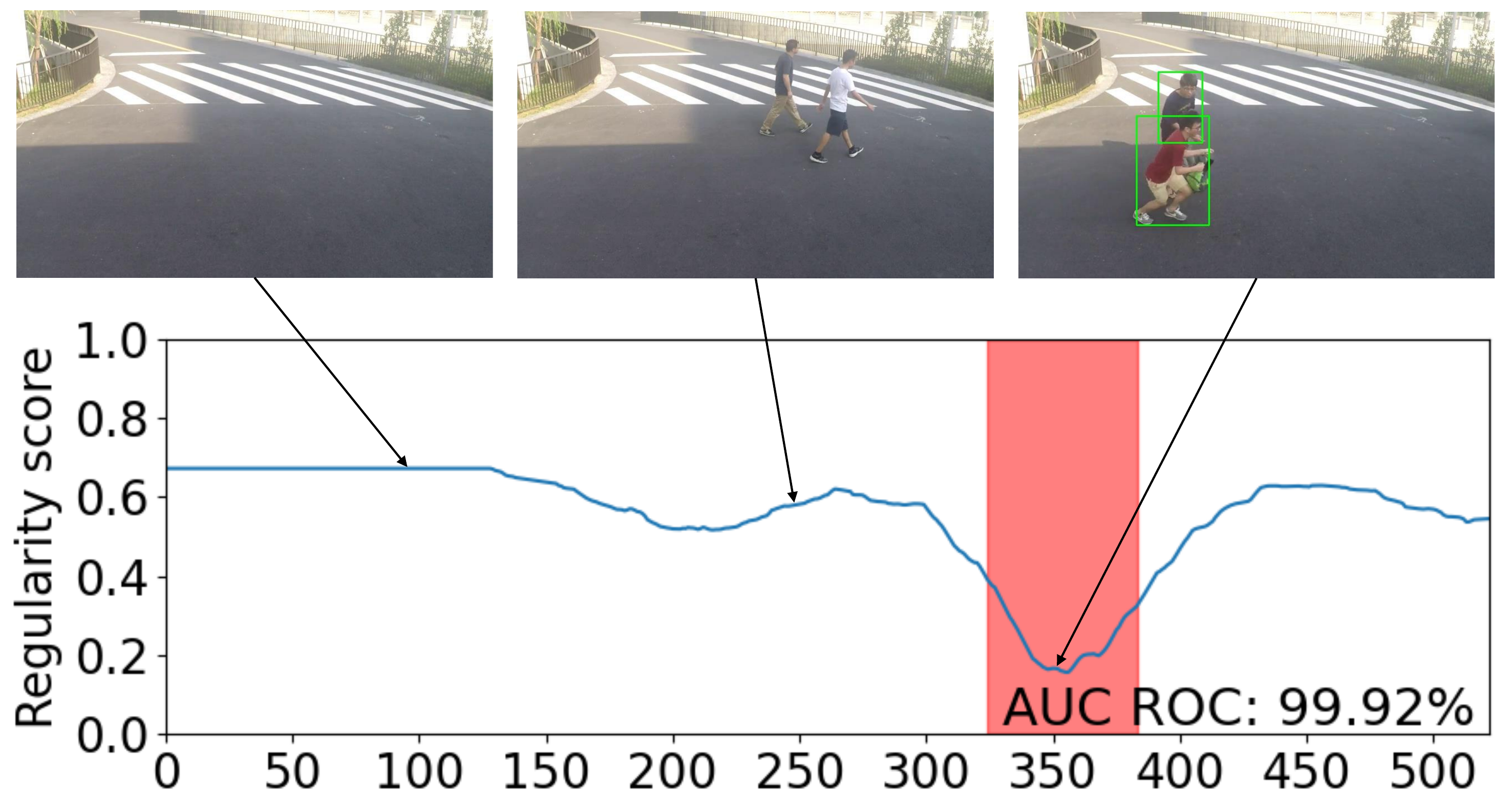}}
	\subfloat[Test video 12\_0149 from STC.]{
		\includegraphics[width=0.49\textwidth]{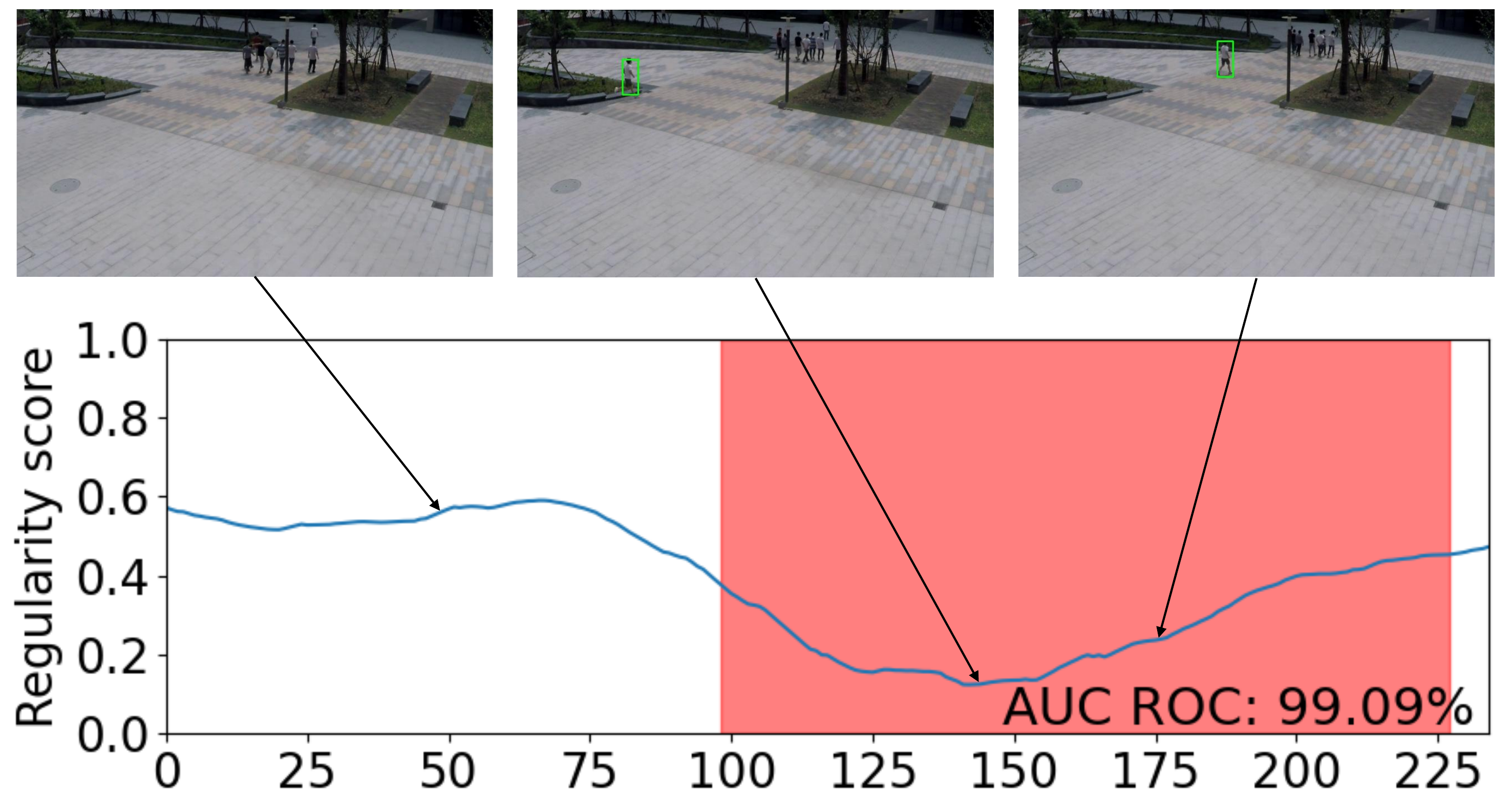}}
	
	\caption{Regularity score curves by our method on STC. The light red shaded regions represent the ground-truth segments of abnormal events.} 
	\label{fig:stc}
\end{figure}

\section{Visual Results}

We provide visual results on UCSD Ped2 \cite{mahadevan2010anomaly}, CUHK Avenue \cite{lu2013abnormal} and STC \cite{luo2017revisit}, shown in Figure \ref{fig:ped2}, Figure \ref{fig:avenue} and Figure \ref{fig:stc}, respectively. Clearly, the regularity scores correlate strongly with the ground-truth temporal segments of the abnormal events, indicating the effectiveness of our method.

\section{Classification \emph{vs.} Regression}
We first convert each position label to one-hot format and then use mean square error (MSE) to regress entries of the one-hot label for each frame/patch. We obtain 83.9\% on STC \emph{vs.} 84.3\% (ours), showing that multi-label formulation is robust to different losses.

\section{Action Recognition}
Both action recognition and VAD require learning spatio-temporal features for classification. But the features they require are different - VAD expects the features sensitive to more subtle changes leading to higher discrimination, while our pretext task also benefits action recognition (as shown by the preliminary results under the fine-tuning protocol on UCF-101 \cite{soomro2012ucf101} in Table \ref{tab:ucf}).

\begin{table}[!h]
	\caption{Results on UCF-101.}
	\label{tab:ucf}
	\centering
	\begin{tabular}{c|c}
		\hline
		\textbf{Method} & \textbf{Accuracy}  \\
		\hline\hline
		Shuffle \& Learn \cite{misra2016shuffle} & 50.2 \\
		OPN \cite{lee2017unsupervised} & 56.3 \\
		VCOP \cite{xu2019self} &  64.9  \\
		Ours  &  \textbf{67.7}  \\  \hline
	\end{tabular}
\end{table}

\end{document}